%% file: main.tex
\documentclass[letterpaper, 10pt, conference]{ieeeconf}

\IEEEoverridecommandlockouts                              % This command is only needed if 
\input{preamble}

\input{notation}
\begin{document}
% \pagenumbering{gobble} % suppress page numbering

\title{\LARGE \bf{Beyond the Current Scene: \\Event-Referential Grasping with Active View Selection}
\vspace{-0.3cm}
}

% \author{Anonymous Author(s)}
\author{
Hyunjoon Lee\textsuperscript{1*}\quad
Haebeom Jung\textsuperscript{1*}\quad
Eunsung Cha\textsuperscript{1}\quad
Daeun Lee\textsuperscript{1}\quad
Yu-Chiang Frank Wang\textsuperscript{2} \\
Jaesung Choe\textsuperscript{2}\quad
Jaesik Park\textsuperscript{1$\dagger$} \\
\textsuperscript{1}Seoul National University\quad
\textsuperscript{2}NVIDIA
\vspace{-0.6cm}%
\thanks{$^*$Equal contribution, $^\dagger$Corresponding author}%
\thanks{
\textbf{Links:} \href{https://www.haebeom.com/BeyondCSe}{Project page}
$\mid$ \href{https://github.com/SNU-VGILab/BeyondCSe}{Github code}
}%
}

\newcommand{\Ours}{\textbf{BeyondCSe}}
%%%%%%%%%%%%%%%%%%%%%%%%%%%%%%%%%%%%%%%%%%%%%%%%%%%%%%%%%%%%%%%%%%%%%%%%%%%%%%%%
% Teaser: placed directly below the title and above the abstract.
% \IEEEaftertitletext injects full-width, non-floating material into the title block,
% so it cannot drift away like a figure* float would.

\maketitle

\input{figures/teaser}

\begin{abstract}
%Event-referential grasping task requires identifying a target object or part from event history rather than from the current observation alone.
%We present \ours{}, a zero-shot system that uses video reasoning to identify the requested object or part and directly grasps it when visible.
%When the target is not visible, Event-Conditioned Active Perception combines its recovered 3D observations with initial scene geometry to form a volumetric belief over its location.
%We combine this belief with visibility uncertainty to evaluate how much each viewpoint can narrow down possible target locations, updating both the belief and scene map as new observations arrive.
%Real-robot experiments using a single wrist-mounted RGB-D camera yield {\color{red}\textbf{\{TBU: grasping results and baseline comparisons\}}} under visible and occluded conditions.
%Search and ablation experiments show {\color{red}\textbf{\{TBU: effects of event history and view selection\}}}.
A robot that observes people interacting with objects should be able to carry out later requests that refer back to those interactions.
Such requests may specify a grasp target by the role it played in a past event rather than by its name or appearance.
Moreover, the target may no longer be visible when the robot is asked to act.
We present \Ours{}, a zero-shot robotic grasping system for this event-referential setting.
Given the event history and the current scene, the system identifies the requested object or part and localizes it for grasping.
If the target is occluded, it combines an event prior recovered from the history with current scene geometry to select camera viewpoints likely to reveal the target.
The system uses pretrained models without additional task-specific training.
%Real-robot experiments using a single wrist-mounted RGB-D camera achieve grasp success rates of 76\% and 77\% for initially visible and occluded targets, respectively, exceeding the strongest baseline in each condition by 36 and 22 percentage points.
In real-robot experiments with a single wrist-mounted RGB-D camera, it achieves grasp success rates of 76\% and 77\% for initially visible and occluded targets, respectively, compared with 40\% and 55\% for the strongest baseline in each condition.
On four additional scenes with heavy occlusion, it increases grasp success rates from 75\% to 95\% while reducing the mean number of views from 3.35 to 2.20, compared with an active-perception baseline given the target's ground-truth 3D bounding box.
\end{abstract}

%%%%%%%%%%%%%%%%%%%%%%%%%%%%%%%%%%%%%%%%%%%%%%%%%%%%%%%%%%%%%%%%%%%%%%%%%%%%%%%%

\IEEEpeerreviewmaketitle

%%%%%%%%%%%%%%%%%%%%%%%%%%%%%%%%%%%%%%%%%%%%%%%%%%%%%%%%%%%%%%%%%%%%%%%%%%%%%%%%

\input{sections/1_intro_v3}

\input{sections/2_related_work}

\input{sections/4_method}

\input{sections/5_experiments}

\input{sections/6_conclusion}

\addtolength{\textheight}{-0cm}   % This command serves to balance the column lengths

%%%%%%%%%%%%%%%%%%%%%%%%%%%%%%%%%%%%%%%%%%%%%%%%%%%%%%%%%%%%%%%%%%%%%%%%%%%%%%%%

%%%%%%%%%%%%%%%%%%%%%%%%%%%%%%%%%%%%%%%%%%%%%%%%%%%%%%%%%%%%%%%%%%%%%%%%%%%%%%%%


% \section*{ACKNOWLEDGMENT}

%%%%%%%%%%%%%%%%%%%%%%%%%%%%%%%%%%%%%%%%%%%%%%%%%%%%%%%%%%%%%%%%%%%%%%%%%%%%%%%%
{\small
    \bibliographystyle{IEEEtran}
    \bibliography{reference}
}

\end{document}

%% file: preamble.tex
\usepackage{amsmath,amsfonts}
\usepackage{algorithmic}
\usepackage{array}
\usepackage{subfig}
\usepackage{textcomp}
\usepackage{stfloats}
\usepackage{url}
\usepackage{verbatim}
\usepackage{graphicx}
\usepackage{capt-of}
\usepackage{balance}
\usepackage{cuted}

\usepackage{multirow}
\usepackage{booktabs}
\usepackage[dvipsnames]{xcolor}
\usepackage{colortbl}

\colorlet{graspRankFirst}{Green!25}
\colorlet{graspRankSecond}{SpringGreen!45}
\colorlet{graspRankThird}{Yellow!30}

\usepackage{hyperref}
\usepackage[capitalise,noabbrev]{cleveref}

\definecolor{colorf}{HTML}{FFB2B2}
\definecolor{colors}{HTML}{FFD9B2}
\definecolor{colort}{HTML}{FFFFB2}
\definecolor{colorref}{HTML}{C8C8C8}

\newcommand{\etal}{\textit{et al.}}

%% file: figures/teaser.tex
% \IEEEaftertitletext{%
%   \vspace{0.5\baselineskip}%
%   \noindent\begin{minipage}{\textwidth}
%     \centering 
%     \includegraphics[width=0.95\textwidth]{assets/teaser_final.pdf} 
%     \captionof{figure}{\textbf{Event-referential grasping.}
%     %
%     %In the history video, the \textit{drill} moved first and the \textit{can} moved last.
%     In the history video, the \textit{drill} and the \textit{can} were the first and last objects moved, respectively.
%     (A) VoLo~\cite{volo_2026}, a VLM orchestrator, correctly identifies the target of the instruction and passes its text description to either a VLA ($\pi_{0.5}$~\cite{pi05}) or a WAM (Cosmos3-Nano-Policy~\cite{cosmos3}).
%     % Given video from the recorded interaction, VoLo~\cite{volo_2026}, a recent VLM orchestrator, identifies the target and delegates grasping to two representative policies: the VLA $\pi_{0.5}$~\cite{pi05} and the WAM Cosmos3-Nano-Policy~\cite{cosmos3}.
%     Nevertheless, both downstream models fail to locate and grasp the occluded target.
%     (B) Our system instead initializes a spatial belief from past 3D target observations, actively selects a viewpoint that reveals the target, and successfully grasps it.
%     }
%     \label{fig:teaser}
%   \end{minipage}%
%   \vspace{0.5\baselineskip}%
% }
\begin{strip}
    \centering
    \includegraphics[width=0.95\textwidth]{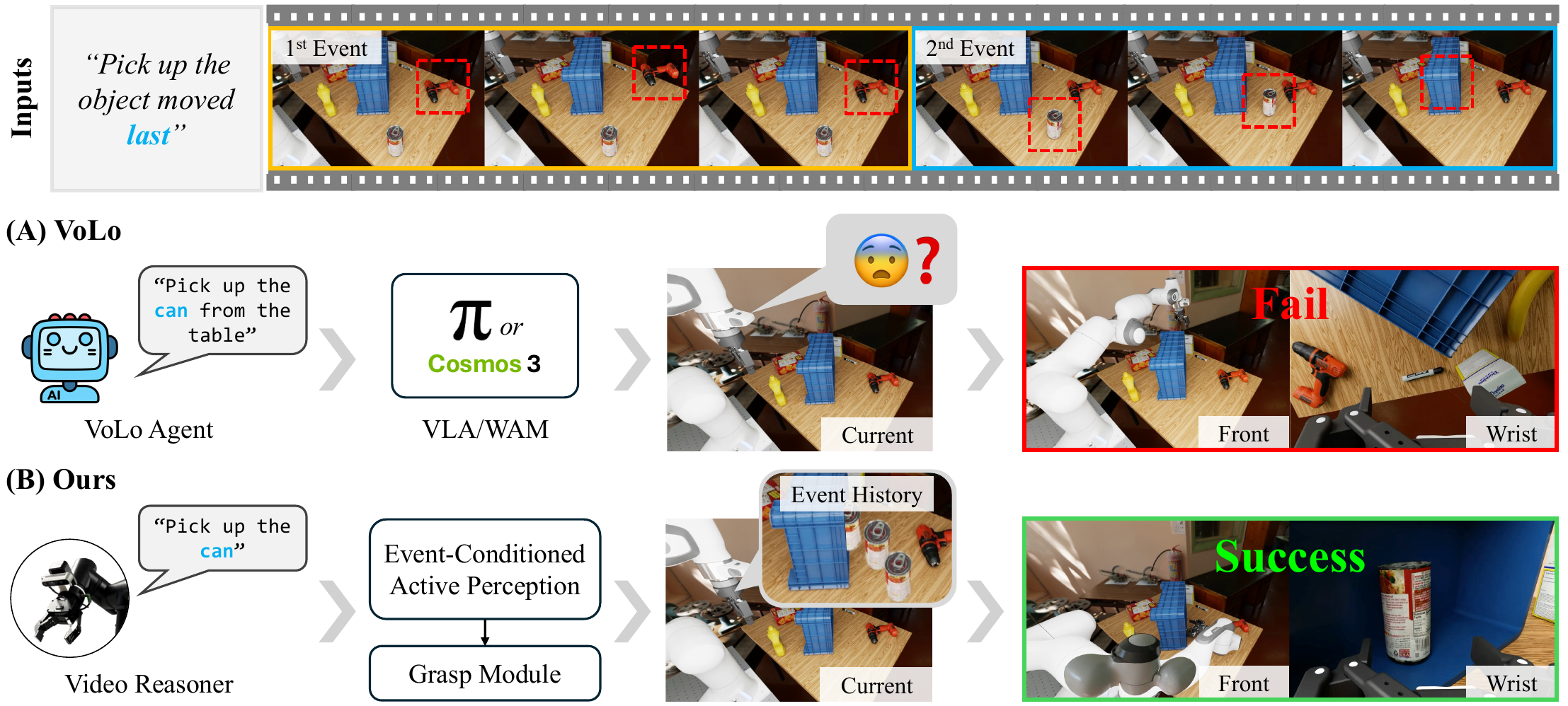}
    \captionof{figure}{\textbf{Event-referential grasping.}   
    %In the history video, the \textit{drill} and the \textit{can} were the first and last objects moved, respectively.
    (A) VoLo~\cite{volo_2026}, a VLM orchestrator, correctly identifies the target of the instruction and passes its text description to either a VLA ($\pi_{0.5}$~\cite{pi05}) or a WAM (Cosmos3-Nano-Policy~\cite{cosmos3}).
    % Given video from the recorded interaction, VoLo~\cite{volo_2026}, a recent VLM orchestrator, identifies the target and delegates grasping to two representative policies: the VLA $\pi_{0.5}$~\cite{pi05} and the WAM Cosmos3-Nano-Policy~\cite{cosmos3}.
    Nevertheless, both downstream models fail to ground and grasp the occluded target.
    % (B) Our system instead initializes a spatial belief from past 3D target observations, actively selects a viewpoint that reveals the target, and successfully grasps it.
    (B) Our system instead initializes a spatial belief from past observations, actively selects a viewpoint that reveals the target, and successfully grasps it.
    }
    \label{fig:teaser}
\end{strip}

%% file: sections/1_intro_v3.tex
\section{Introduction}
\label{sec:intro}

% A robot working alongside people must operate in a workspace that is continually reshaped as objects are moved, used, and put away.
% After such an interaction, a person may ask the robot, \textit{``Pick up the object I just used.''}
% Such an event-referential instruction specifies the target by its role in a past event rather than by its visual attributes in the current scene.
Real-world workspaces have histories: before a robot is asked to act, objects may have been moved, used, and put away.
An instruction can refer to this history, as in \textit{``Pick up the object I just used.''}
We call such an instruction \emph{event-referential}: it identifies a target object or part by its role in a past event rather than by attributes in the current scene.
The robot must therefore resolve the reference to the correct instance using the event history.
% If the interaction leaves the target occluded, the history must provide not only its identity but also spatial evidence of where it went.
If the target is no longer visible, the history must provide not only its identity but also spatial evidence of where it went.
The robot can then use this evidence to reach a viewpoint from which the target is visible and graspable.
% In our analysis, ...

Language-guided manipulation methods map an instruction and the robot's current observations to a grasp~\cite{pi05,cosmos3,lerf_togo_2023,graspsplat_2025,point2act_2025,graspmolmo_2025}.
These methods typically resolve targets described by name, category, appearance, or part within the current scene.
A target defined by its role in a past event, however, cannot be identified from the current scene alone.
Recent systems incorporate event history by conditioning actions on past observations~\cite{memoryvla_2026,memer_2026} or by passing a reasoned target description to a policy~\cite{volo_2026}.
These approaches can recover the identity of a target that is no longer visible, but target identification alone does not provide the current visual evidence needed for grasping.
Our method addresses this gap by using the event history to determine both the intended object and where the camera should move to see it.
As shown in \Cref{fig:teaser}, VoLo~\cite{volo_2026} correctly identifies the target and delegates grasping to either a vision-language-action~(VLA) model~\cite{pi05} or a world action model~(WAM)~\cite{cosmos3}.
Yet neither downstream model can grasp the target while it remains occluded.
Our system instead selects a viewpoint that reveals the target before grasping it.

Active perception supplies this missing visual evidence by selecting new viewpoints likely to reveal a hidden target.
Prior work derives view scores from the current scene using geometric information gain~\cite{breyer_2022}, predicted grasp affordance~\cite{acenbv_2023}, or object--occluder relations when the target has never been seen~\cite{visograsp_2025}.
The current scene alone, however, provides limited guidance about an object displaced during an earlier event.
Our event-conditioned prior instead uses the target's own past observations to estimate where it went.
Our view-selection objective further models the visibility of likely target regions, discounting candidate views whose rays pass through unobserved space rather than treating that space as free.

%To this end, we present \Ours{}, a zero-shot grasping system that uses event history to search actively for a viewpoint from which the target can be grasped.
% To this end, we present \Ours{}, a zero-shot grasping system that locates the target of an event-referential instruction by combining the event history with the current scene, and actively searches for a viewpoint when the target is not visible.
To this end, we present \Ours{}, a zero-shot grasping system that grounds event-referential targets using the event history and current scene, actively seeking new viewpoints when necessary.
The video reasoning first associates the instruction with the target in the event history and then attempts to ground it in the current observation.
If grounding succeeds, the module passes an action point to the grasping backend. Otherwise, it passes earlier target observations to event-conditioned active perception.
These observations are combined with current scene geometry to initialize a volumetric target belief.
The active-perception module scores candidate viewpoints by combining the target belief with transmittance-aware visibility and updates the belief after each new observation until the target is found or the sensing budget is exhausted.

\noindent Our main contributions are as follows:
\begin{itemize}
    % \item A zero-shot grasping pipeline that identifies objects or parts specified by event-referential instructions and recovers an event prior using an off-the-shelf multimodal large language model (MLLM) and point tracker.
    \item A zero-shot grasping pipeline that grounds event-referential objects or parts and recovers an event prior with an off-the-shelf multimodal large language model (MLLM) and point tracker.
    \item A probabilistic volumetric formulation integrating an event-conditioned spatial prior with transmittance-aware visibility for Bayesian belief updates and view selection.
    % \item A real-robot evaluation using a wrist-mounted RGB-D camera, comparing grasping and active-perception methods, and examining how evidence from the event history affects search cost and grasping success.
    \item A real-robot evaluation of grasping and active perception with a wrist-mounted RGB-D camera, including the effects of event-history evidence on search cost and grasping success.
\end{itemize}

%% file: sections/2_related_work.tex
\section{Related Work}
\label{sec:related_work}

% \subsection{Language-Guided Grasping and Grounding}
% \myparagraph{Vision-language action for manipulation}
\subsection{Language-Guided Grasping and Grounding}
LERF-TOGO~\cite{lerf_togo_2023} and GraspSplats~\cite{graspsplat_2025} ground language queries in 3D representations built from multiple views.
For point-based grounding, GraspMolmo~\cite{graspmolmo_2025} adapts a point-grounding MLLM to predict a task-oriented grasp point from a single image, while Point2Act~\cite{point2act_2025} distills multi-view MLLM point predictions into a 3D relevancy field.
At the system level, VoLo~\cite{volo_2026} coordinates VLAs, perception models, and action primitives for long-horizon manipulation, including tasks involving memory and complex references.
% Our focus is identifying a target from a past event and using its observed history to guide search when it is occluded.
Our setting goes beyond current scene grounding by identifying an object or part through its role in an event history and using its event prior to guide search under occlusion.

% \subsection{History-Dependent Manipulation and Episodic Memory}
% \myparagraph{History-dependent manipulation}
\subsection{History-Dependent Manipulation and Episodic Memory}
EgoLoc~\cite{egoloc_2023} recovers the past 3D location of an object specified by an image query, while VideoAgent~\cite{videoagent_2024} retrieves frames to answer questions about a video.
For robot navigation, ReMEmbR~\cite{remembr_2025} queries a robot's spatio-temporal memory to generate navigation goals.
For manipulation, MemoryVLA~\cite{memoryvla_2026} incorporates past observations into action generation, and MemER~\cite{memer_2026} selects relevant keyframes to generate instructions for a low-level policy.
Complementing these methods, RoboMME~\cite{robomme_2026} systematically evaluates history-dependent manipulation, including ordinal references and memory of temporarily hidden objects.
% For target search, our system recovers past 3D observations of the referenced target without additional task-specific training.
Unlike memory-conditioned policies, our system uses off-the-shelf models to recover explicit 3D observations of a target specified by an event-referential instruction without task-specific training.

% \subsection{Active Perception and Occluded-Target Search}
% View planning for grasping uses geometric information gain~\cite{breyer_2022} or predicted grasp affordance~\cite{acenbv_2023}.
% ActiveGrasp~\cite{activegrasp_2026} estimates view information gain from calibrated grasp-success uncertainty.
% Under severe occlusion, VISO-Grasp~\cite{visograsp_2025} uses vision-language reasoning for active view planning and grasping, including fully invisible targets.
% At the policy level, SaPaVe~\cite{sapave_2026} and ActiveVLA~\cite{activevla_2026} integrate active perception with learned manipulation policies.
% In mechanical search, SMS~\cite{sms_2023} supplies an LLM-derived semantic occupancy distribution to guide object movements that expose a hidden target.
% Our system identifies the target from a recorded event and initializes a search prior from its last observation and observed trajectory.
% We evaluate how this target-specific history guides view selection.
% Existing methods select views using observed geometry, predicted grasp affordance or grasp uncertainty, or object--occluder relations inferred from the current scene, while SMS uses a category-level semantic occupancy prior.
% In contrast, our system constructs a target-specific spatial belief from the target's recorded observations and trajectory, refines it as new observations arrive, and selects views that best examine likely target locations without assuming that unobserved space is empty.

% \myparagraph{Active perception for occluded object search}
\subsection{Active Perception for Occluded-Target Search}
Active view planning for grasping selects viewpoints based on geometric information gain~\cite{breyer_2022}, predicted grasp affordance~\cite{acenbv_2023}, or uncertainty in a calibrated grasp-success model~\cite{activegrasp_2026}.
For severe occlusion, VISO-Grasp~\cite{visograsp_2025} reasons about object--occluder relations in the current scene to adjust the viewpoint or remove an inferred occluder.
Learned approaches couple active perception with manipulation: SaPaVe~\cite{sapave_2026} predicts camera and manipulation actions, while ActiveVLA~\cite{activevla_2026} selects virtual views rendered from reconstructed 3D input.
% SMS~\cite{sms_2023} addresses mechanical search by using category-level semantic associations to predict where a hidden target may lie and to guide object removal.
These methods derive search guidance from current observations, grasp predictions, or generic semantic associations.
% Our system instead turns the referenced target's event history into an instance-specific belief, revises it with each observation, and selects views by the visibility of likely target locations while discounting rays through unobserved space.
Our system instead derives an instance-specific belief from the target's event history and updates this belief after each observation.
The system then selects viewpoints based on the visibility of likely target locations, discounting rays through unobserved space.

%% file: sections/4_method.tex
\section{Method}

\subsection{Problem Formulation}
\label{sec:problem}
% jchoe
%A wrist-mounted RGB-D camera remains stationary while recording a person's tabletop event history as a video $\mathcal{V}$. The instruction $\ell$ arrives after the observed events end. The final RGB-D frame serves as the robot's initial observation $(I_0,D_0)$ for executing the instruction. The camera intrinsics and camera-to-base transform are known. The instruction identifies an object or part through a past event. The target may be indistinguishable from other objects in the initial image or occluded from the initial view.
%Our goal is to obtain a 3D action point $p^{\star}$ in the robot base frame using the event history and initial observation, with additional observations acquired as needed. We index robot observations by $t$, with $t=0$ denoting the initial observation and $(I_t, D_t)$ denoting the current observation at step $t$.

%Given an event history video~$\mathcal{V}$, a language instruction~$\ell$, and a current RGB-D observation~$(I_t,D_t)$ at time $t$, our model aims to predict a 3D action point~$p^{\star}$.The camera is stationary and camera parameters are known. Note that we index current observations by $t$ and $t=0$ denotes an initial observation. The instruction identifies an object or part through a past event. The target may be indistinguishable from other objects in the initial image or occluded from the initial view.
Given an RGB-D video~$\mathcal{V}$ of a person's tabletop event history and a language instruction~$\ell$ issued after the events end, our system aims to obtain a 3D action point~$p^{\star}$ in the robot base frame.
The video is recorded by a wrist-mounted camera that remains stationary during recording, and its final frame serves as the robot's initial observation~$(I_0,D_0)$.
The robot may then move the camera to acquire additional observations~$(I_t,D_t)$, indexed by step~$t$, with camera intrinsics and camera-to-base transform known at every step.
The instruction identifies an object or part through a past event.
The target may be indistinguishable from other objects in the initial image or occluded from the initial view.

\input{figures/overview_pipeline}

\subsection{System Overview}
\label{sec:system_overview}
The two modules in Fig.~\ref{fig:overview} are connected through the target description and spatial observations.
If video reasoning (Sec.~\ref{sec:video_reasoning}) obtains a valid 3D action point from the initial observation, the system passes it to grasp validation, skipping target-location search.
Otherwise, recovered historical 3D observations are passed to event-conditioned active perception (Sec.~\ref{sec:nbv}) to initialize a target belief together with geometry from the initial observation.
New observations support target pointing and update the map and belief; once the target is confirmed, the system proceeds to grasp validation.
The pipeline uses an off-the-shelf MLLM~\cite{qwen3_2025} and point tracker~\cite{cotracker3_2025} without additional task-specific training.

\subsection{Video Reasoning}
\label{sec:video_reasoning}
\input{figures/reasoning}
\subsubsection{Selecting the referent}
Figure~\ref{fig:reasoning}(A) shows the inputs and outputs of the four stages.
\textsc{Record} receives the video without the instruction and forms a time-ordered event record $\mathcal{E}=(e_1,\ldots,e_N)$, where $N$ is the number of recorded events.
Each event $e_i=(a_i,o_i,r_i)$ describes an action, its object, and another involved object, if any.
\textsc{Select} reads this record and the instruction to produce an appearance description of the requested object or part (\emph{target}) and an event description that distinguishes it (\emph{cue}).
The target is passed to \textsc{Point}, and the cue to \textsc{Locate}.

\subsubsection{Grounding in the initial image}
\textsc{Locate} uses the video and cue to propose a box in $I_0$.
\textsc{Point} receives the target and the corresponding crop, and returns an action pixel.
The crop narrows the search region and enlarges small parts, as illustrated by the example in Fig.~\ref{fig:reasoning}(B).
If crop-based pointing returns no usable point, the search region is expanded; if no usable box is available, the full image is used.
The returned pixel is mapped to the full initial image $I_0$ and lifted to $p^{\star}$ using valid depth and the camera pose.

\subsubsection{Recovering historical locations}
When pointing in the initial image returns no usable point, we search sampled past frames from recent to earlier ones for pixels matching the target.
We track these candidates together and reject trajectories that remain observed at the end of the video.
The remaining candidates are checked for appearance, most recent first, and the first to pass is selected.
We lift its visible track samples with valid depth into the robot base frame to obtain
\begin{equation}
  \mathcal{P}=(q_1,\ldots,q_m),
  \label{eq:prior}
\end{equation}
Here, $m$ is the number of valid 3D observations, and $q_m$ is the last one.
The 3D event track $\mathcal{P}$ initializes the target belief in Sec.~\ref{sec:nbv}.

\subsection{Event-Conditioned Active Perception}
\label{sec:nbv}

\input{figures/avs_pipeline.tex}

Figure~\ref{fig:avs_pipeline} summarizes the loop: at $t{=}0$, the \emph{event history} and \emph{current observation} initialize the \emph{target belief} and \emph{map}.
Each view acquired by \emph{belief-guided view search} then provides the next observation, which updates the map and belief.
Once the target is confirmed, grasp validation either returns a plan or requests a target-centered refinement view.

\subsubsection{Event-Conditioned Belief Initialization}
We construct the initial map $\mathcal M_0$ as a Truncated Signed Distance Function (TSDF) from the initial RGB-D observation $(I_0,D_0)$ and define $\Omega\subset\mathbb R^3$ as the geometrically admissible workspace, retaining regions unresolved by occlusion or missing depth.
The map represents observed geometry, whereas $b_t(x)$ is the target-location probability density over $\Omega$ after observation step $t$ and integrates to one.
The stationary target has no intervening motion model.
We initialize this belief from the 3D event track $\mathcal P$ in Sec.~\ref{sec:video_reasoning}, without requiring a current target mask, bounding box, or center.

Let $\gamma:[0,L]\to\mathbb R^3$ be the piecewise-linear path through $\mathcal P$, parameterized by arc length, with $\gamma(L)=q_m$ and total length $L=\sum_{i=2}^{m}\|q_i-q_{i-1}\|_2$.
Using the terminal segment of length $h=L/\rho$ for $\rho\geq1$, we set
\begin{align}
\mu_e&=q_m,\qquad \Delta(s)=\gamma(s)-q_m,\\
C_{\mathrm{tail}}&=\frac{1}{h}\int_{L-h}^{L}\Delta(s)\Delta(s)^\top\,ds,\\
\Sigma_e&=\delta_0^2\mathbf I_3+C_{\mathrm{tail}}.
\label{eq:event-belief}
\end{align}
Thus, the prior is centered at the last observation and shaped by the terminal path without extrapolating it.
Here, $\mathbf I_3$ is the $3\times3$ identity matrix, and the subscript $e$ denotes the event-conditioned prior.
The parameter $\delta_0>0$ sets the minimum spread, and $C_{\mathrm{tail}}=0$ for a single observation or zero-length path.
With $\mathcal N_{\Omega}$ denoting the Gaussian restricted and normalized over $\Omega$, we use
\begin{equation}
b_0(x)=(1-\epsilon)\cdot\mathcal N_{\Omega}(x;\mu_e,\Sigma_e)
+\epsilon\cdot\frac{\mathbf{1}_{\Omega}(x)}{|\Omega|}.
\label{eq:initial-belief}
\end{equation}

Here, $\mathbf{1}_{\Omega}$ denotes the indicator of $\Omega$, and $|\Omega|$ denotes its volume.
The mixture weight $0<\epsilon<1$ assigns nonzero probability throughout $\Omega$, allowing the search to recover when the historical estimate is inaccurate.
Without valid 3D history, $b_0$ is uniform over $\Omega$.

\subsubsection{Belief-Guided View Search}
We generate candidate views around high-probability belief regions and retain the feasible set $\Xi_t$ after kinematic, self-collision, and scene-collision checks.
For each $\xi\in\Xi_t$, we render $\widehat D_t^\xi$ from $\mathcal M_t$ and evaluate the negative-observation likelihood $\mathcal{L}^-_t(x;\xi)\in(0,1]$ at each target-location hypothesis $x\in\Omega$.
This look-ahead likelihood uses the same depth-consistency and target-miss factors as the subsequent Bayesian update.

When a ray is not terminated by an observed surface, its visibility confidence is attenuated according to the distance traversed through unobserved space:
\begin{equation}
T_t(x;\xi)=\kappa+(1-\kappa)\exp\!\left[-\sigma_t d_{u,t}(x;\xi)\right].
\label{eq:transmittance}
\end{equation}
Here, $d_{u,t}(x;\xi)\geq0$ is the ray length through unobserved space, $\sigma_t\geq0$ controls attenuation, and $0<\kappa\leq1$ sets a nonzero transmittance floor.
Inspired by the accumulated volumetric transmittance used in NeRF~\cite{nerf_2021}, we use $T_t$ without learning a radiance field: it instead measures confidence in a hypothetical observation through unobserved space.
Together, these components form a coherent probabilistic loop: the event-conditioned spatial prior structures the initial target belief, nonzero observation likelihoods revise it without hard exclusions, and volumetric transmittance discounts views whose apparent informativeness depends on unobserved space.
We score each view by the target-belief mass that a negative observation is expected to downweight:
\begin{align}
\xi_t^{\star}&=\arg\max_{\xi\in\Xi_t} S_t(\xi),\\
S_t(\xi)&=\int_{\Omega}b_t(x)\left[1-\mathcal{L}^-_t(x;\xi)\right]T_t(x;\xi)\,dx.
\label{eq:view-score}
\end{align}
We execute the highest-scoring view that admits a valid motion plan.
\input{figures/task_examples}

\subsubsection{Map and Belief Update}
After executing $\xi_t^\star$, we register the acquired keyframe $(I_{t+1},D_{t+1},\xi_t^\star)$, integrate its depth into $\mathcal M_{t+1}$, and update the same belief scored in Eq.~\eqref{eq:view-score}.
We combine a depth-consistency likelihood $\mathcal{L}_{\mathrm{dep},t+1}(x)$ with a target-miss likelihood $\mathcal{L}_{\mathrm{miss},t+1}(x)$ when $\textsc{Point}$ returns \textsc{Not-Visible}:
\begin{equation}
b_{t+1}(x)\propto b_t(x)\mathcal{L}_{\mathrm{dep},t+1}(x)\mathcal{L}_{\mathrm{miss},t+1}(x).
\label{eq:belief-update}
\end{equation}
The depth likelihood remains neutral wherever the observation provides no valid geometric evidence.
The target-miss likelihood combines predicted visibility with the probability that \textsc{Point} misses a visible target.
Both likelihoods take values in $(0,1]$, so negative observations downweight rather than rule out locations.

At each belief-guided view, $\textsc{Point}$ queries the target description from $\textsc{Select}$.
A lifted pixel yields $p^{\star}$ and ends target search.
Otherwise, the updated posterior is rescored until no feasible view remains or the active view budget is exhausted.

\subsubsection{Target Confirmation and View Refinement}
Once $p^{\star}$ is confirmed, we generate and validate grasp candidates for target association and motion feasibility.
Following Breyer~\etal~\cite{breyer_2022}, incomplete target geometry triggers a feasible target-centered view, map integration, and grasp regeneration.
Unlike their setting with a provided target box, our target region becomes available only after event-conditioned search confirms the target.
The bounded loop terminates with an executable grasp or abstains if none is found within the active-view budget.

%% file: figures/overview_pipeline.tex
\begin{figure*}[t!]
    \centering
    \includegraphics[width=0.92\linewidth]{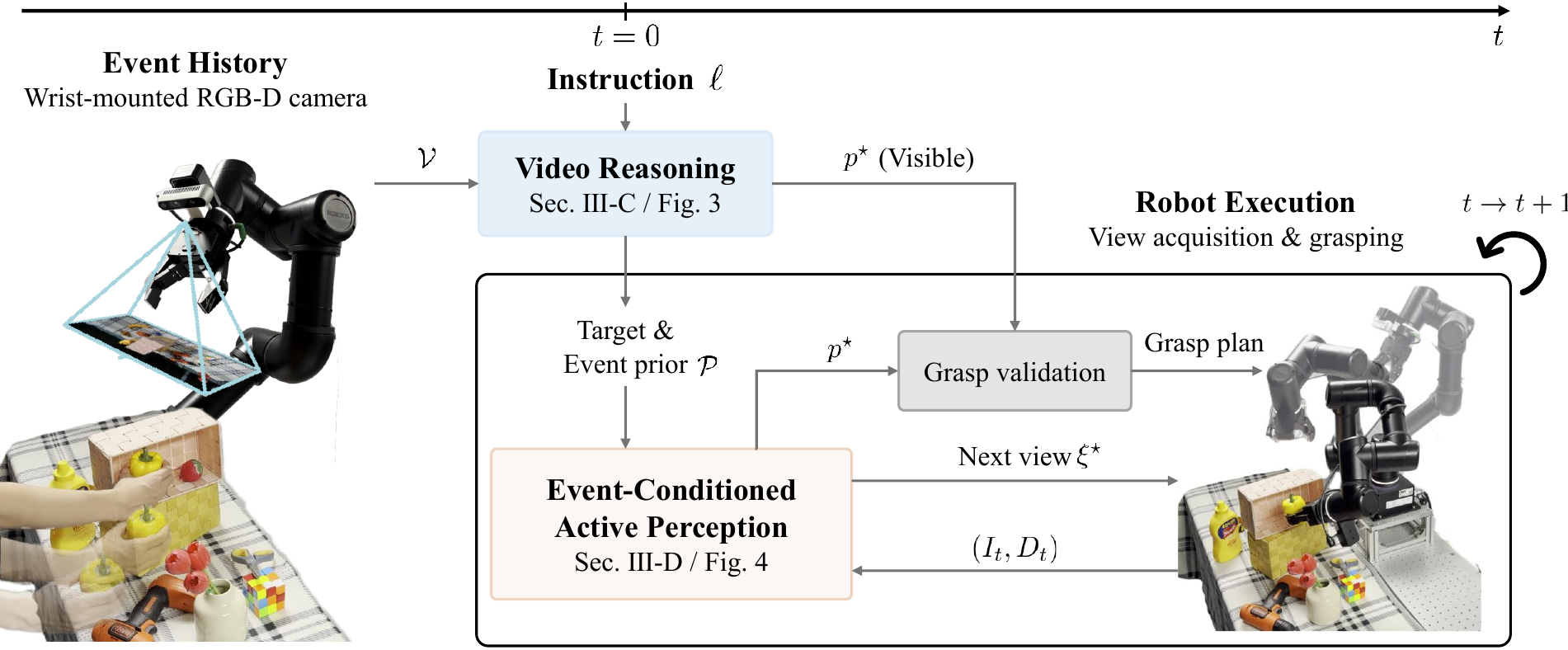}
    \caption{\textbf{Overview of \Ours{}.} 
    Given an event-referential instruction and a recorded video, the video reasoning identifies and grounds the target.
    Historical 3D target observations recovered from the video are denoted by $\mathcal P$.
    If the target is occluded, event-conditioned active perception builds a volumetric belief from prior observations and current geometry, then selects visibility-aware viewpoints until the target can be grasped or search terminates.
    }
    \label{fig:overview}
\end{figure*}

%% file: figures/reasoning.tex
\begin{figure}[t!]
    \centering
    \includegraphics[width=0.99\columnwidth]{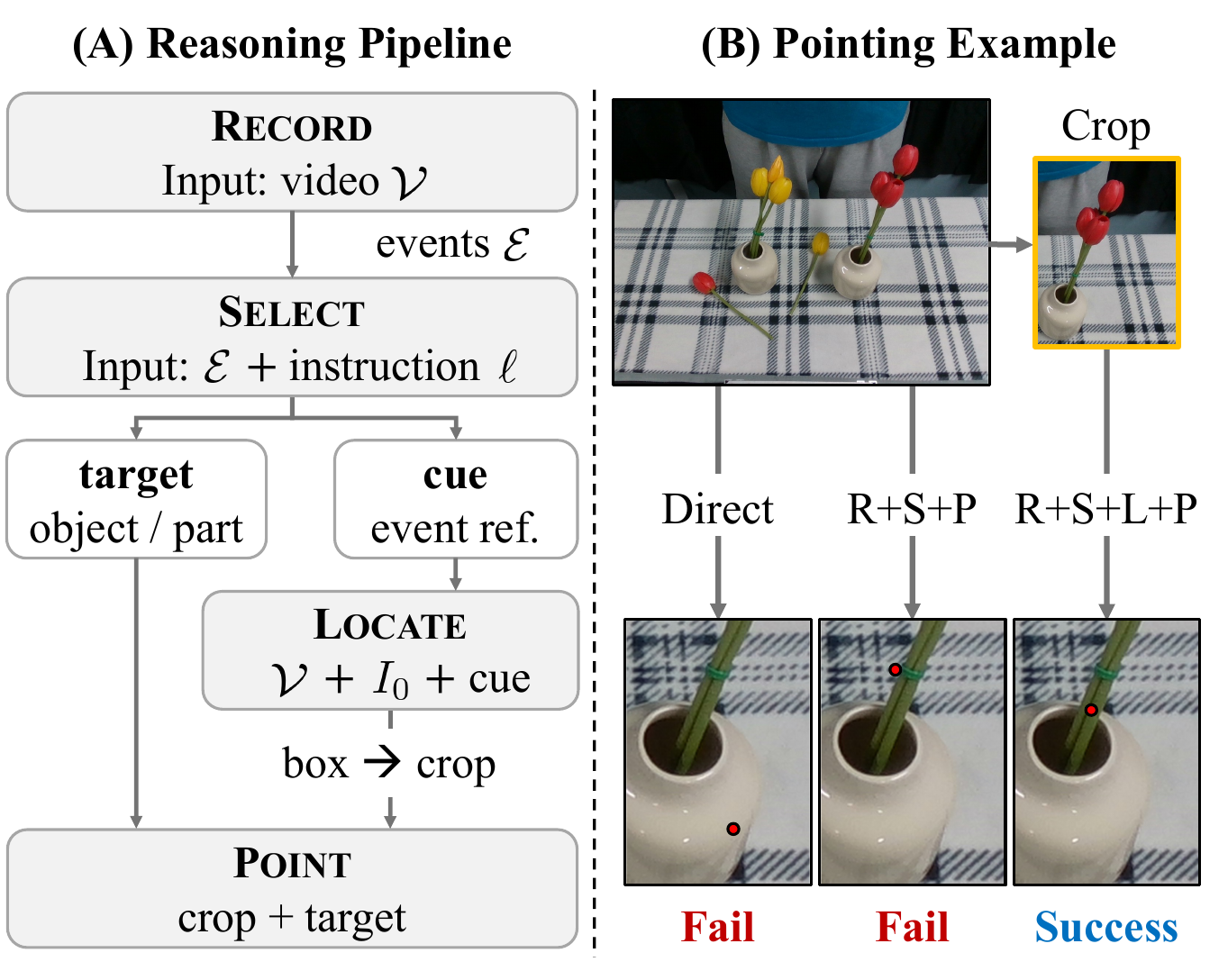}
    \caption{\textbf{Video reasoning.}
    (A) Inputs and outputs of \textsc{Record} (R), \textsc{Select} (S), \textsc{Locate} (L), and \textsc{Point} (P).
    (B) Direct video-to-point, full-image (R+S+P), and crop-based (R+S+L+P) predictions in the same close-up region.
    Red dots indicate predicted results; the latter two configurations share \textsc{Record/Select} outputs.}
    \label{fig:reasoning}
\end{figure}

%% file: figures/avs_pipeline.tex
\begin{figure*}[t!]
    \centering
    \includegraphics[width=0.95\linewidth]{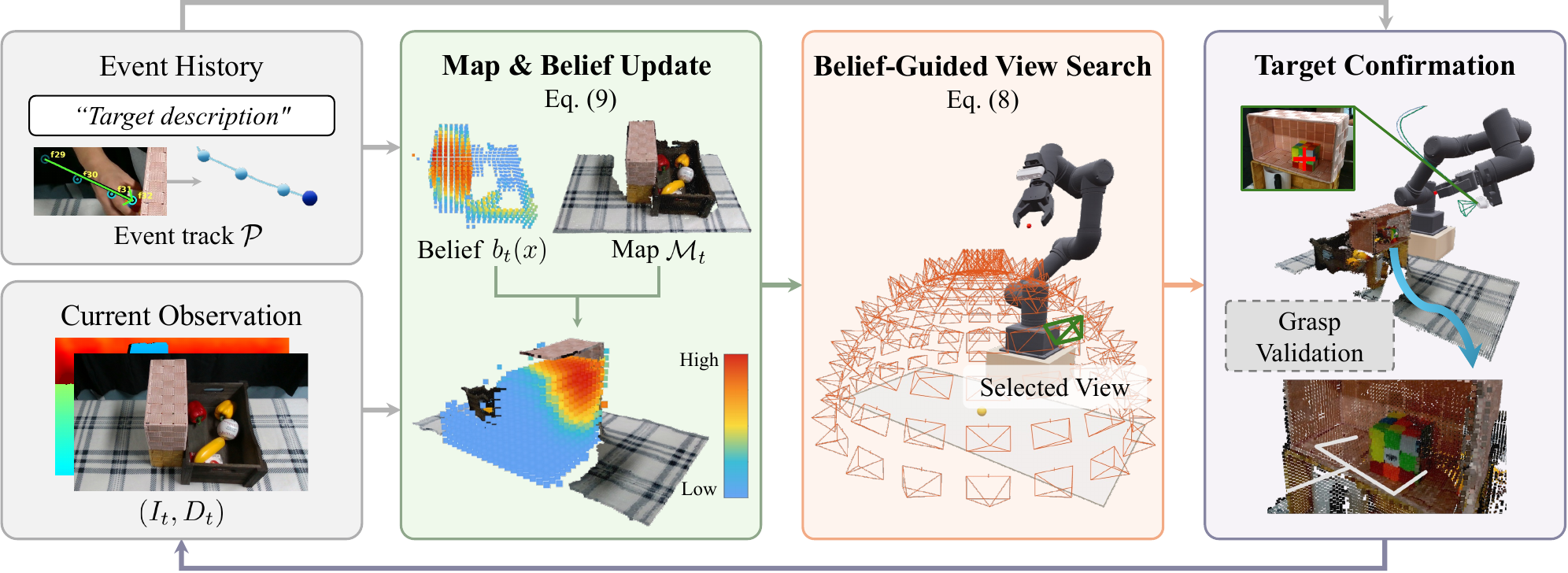}
    \caption{\textbf{Event-conditioned active perception.}
    The event track and initial RGB-D observation initialize the target belief and map.
    Belief-guided view search then acquires physical views, each yielding a current observation that updates both in closed loop.
    After target confirmation, grasp validation requests another view only when the observed geometry is insufficient.
    }
    \label{fig:avs_pipeline}
\end{figure*}

%% file: figures/task_examples.tex
\begin{figure*}[t]
    \centering
    \includegraphics[width=0.95\textwidth]{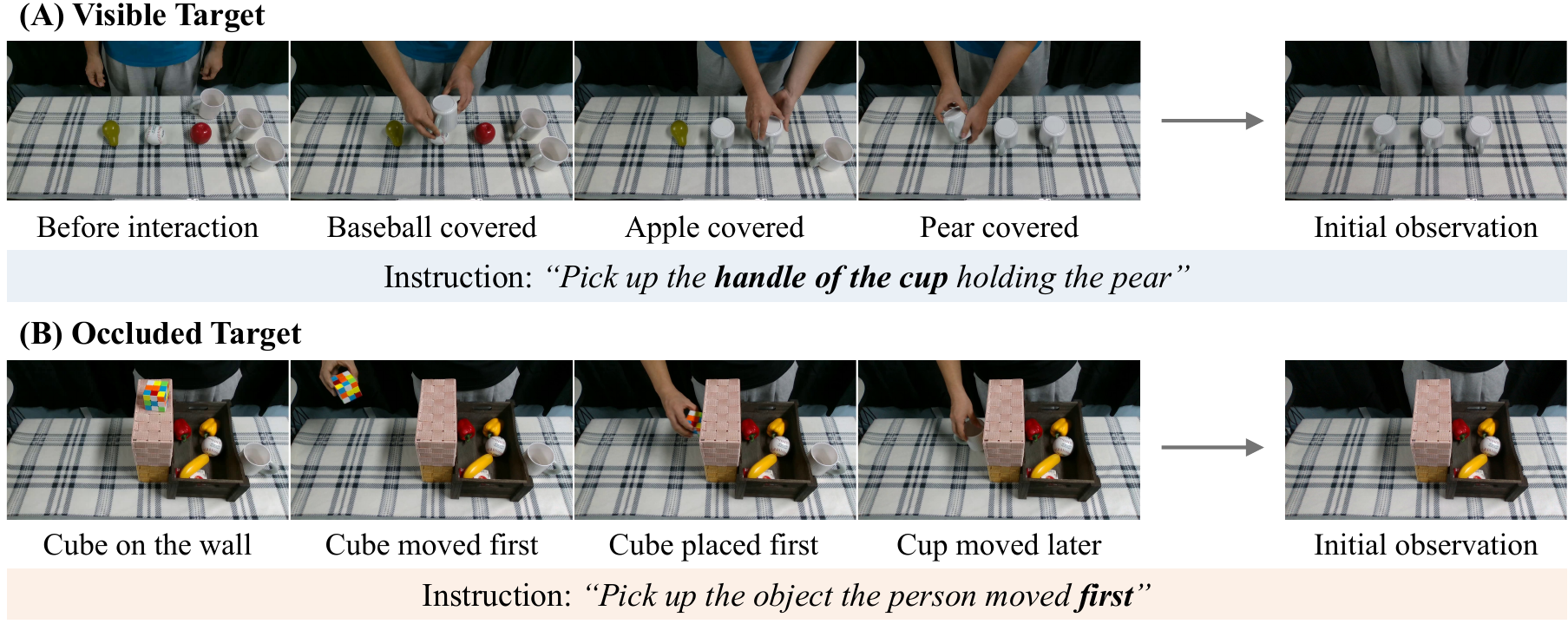}
    \caption{\textbf{Examples of event-referential instructions.}
    Each row shows sampled frames from the event history video, the initial observation for robot execution, and the instruction issued after recording.
    The final video frame serves as the initial observation.
    (A) The requested cup handle is visible, but identifying the cup covering the pear requires the event history.
    (B) The cube is moved first and is occluded by the wall in the initial observation.}
    \label{fig:task_examples}
\end{figure*}

%% file: sections/5_experiments.tex
\section{Experiments}
\label{sec:exp}

% Draft placeholders are visible in the compiled PDF.
\newcommand{\expTBU}[1]{{\color{red}\textbf{\{TBU: #1\}}}}
\subsection{Experimental Setup}

\subsubsection{Hardware Setup}
We conduct all real-world experiments using a ROBOTIS OMY-F3M robot equipped with a wrist-mounted Intel RealSense D435i RGB-D camera.
During video capture, the robot holds the wrist-mounted camera at a fixed observation pose, and the final RGB-D frame becomes the initial observation.
All perception models run on a workstation with a single NVIDIA GeForce RTX 4090 GPU.
All methods share the robot, camera calibration, joint limits, collision checks, and grasp generation and execution.

\subsubsection{Dataset}
Each episode consists of a recorded interaction video, an initial wrist-camera RGB-D observation, and an instruction.
We distinguish visible and occluded conditions by whether the requested object or part is visible in the initial observation.
Our tabletop scenes contain common objects such as produce, cans, mugs, containers, flowers, markers, and toy objects, with other objects serving as potential distractors.
The events include taking objects from containers, placing objects in sequence, and rearranging objects in different temporal orders.
Instructions refer to objects through event order, such as the object moved first or last, and to object parts, such as a flower stem or cup handle.
In the occluded condition, scene occluders hide the queried target from the initial camera view.
Figure~\ref{fig:task_examples} shows example episodes with event-based references and part queries under both conditions.

The visible condition comprises 10 scene--query pairs across four scenes.
The occluded condition comprises 10 scenes with two queries per scene, giving 20 scene--query pairs.
We conduct five real-world trials per scene--query pair, yielding 50 visible and 100 occluded trials, for a total of 150 trials.
These 30 scene--query pairs form the evaluation set for comparison with zero-shot grasping methods.
To compare our event-conditioned active perception module against other active-perception methods, we record four additional heavily occluded scenes.
In these scenes, every queried target is initially out of sight, concealed inside a basket or behind other scene objects, and remains invisible from a fixed bird's-eye view.
Two examples are shown in Fig.~\ref{fig:active_perception_qualitative}(A).
All methods within each comparison use the same scenes and queries.
\input{figures/table_plot}
\input{tables/grasping_overall}

\subsubsection{Implementation Details}
In the \emph{original instruction} condition, grasping baselines receive current scene observations and the original instruction.
In the \emph{reasoned instruction} condition, a separate Qwen3-VL-8B~\cite{qwen3_2025} reasoner converts the history video and instruction into a grounding query.
We generate this query once per episode and instruction and share it across baselines while retaining their visual grounding backbones.
It is generated separately from the pointing instruction in our video reasoning pipeline (Sec.~\ref{sec:video_reasoning}).
Our system uses the same Qwen3-VL-8B for all MLLM stages.
We uniformly sample 96 frames from each event history video and use deterministic decoding with temperature zero.
For Eqs.~\eqref{eq:event-belief} and~\eqref{eq:initial-belief}, we set $\rho=2$, $\delta_0=0.05$~m, and $\epsilon=0.1$.
For Eq.~\eqref{eq:transmittance}, we set $\kappa=0.33$ and estimate $\sigma_t$ from the occupied fraction of voxels newly resolved by the initial depth observation, using $2.0$~m$^{-1}$ when this estimate is unavailable.
A new map keyframe is added after $0.10$~m of camera translation or $15^\circ$ of rotation.
Following the grasp proposal procedure of LERF-TOGO~\cite{lerf_togo_2023}, we generate AnyGrasp~\cite{anygrasp_2023} candidates from virtual views, pool them, and apply non-maximum suppression to duplicate poses.

\subsubsection{Metrics}
We report successful trials out of all trials for 3D localization, planning, and grasping.
%For evaluation, a human annotator marks each queried object or part with a 3D oriented bounding box in the robot base frame.
%The annotation interface presents orthographic top and side projections of the registered multi-view scene together with camera images, from which the annotator specifies the target footprint, in-plane orientation, and height.
%The resulting annotations are withheld from all methods and used only to score localization.
For localization evaluation, a human annotator defines a 3D oriented bounding box for each target object or part in the robot base frame. These annotations are withheld from all methods.
Localization succeeds when the predicted 3D point lies within the corresponding annotated box without an additional distance margin, and planning succeeds when a feasible grasp plan is found for the localized target.
Grasp success measures whether the robot successfully grasps the instructed target.
These metrics reflect successive stages: localization enables planning, and a feasible plan enables grasp execution.
\input{tables/active_perception_ablation}

\subsection{Comparison with Zero-Shot Grasping Methods}
We compare with LERF-TOGO~\cite{lerf_togo_2023}, GraspSplats~\cite{graspsplat_2025}, Point2Act and Point2Act$^\dagger$~\cite{point2act_2025}, and GraspMolmo~\cite{graspmolmo_2025}.
LERF-TOGO and Point2Act use RGB inputs, while GraspSplats, GraspMolmo, and Point2Act$^\dagger$ use RGB-D.
Each baseline is evaluated with both original and reasoned instructions.
Multi-view baselines receive 30 predefined observations covering the scene.
GraspMolmo uses only the initial RGB-D view and is not evaluated when the target is occluded in that view, as indicated by N/A in Fig.~\ref{fig:table_plot}.
Our method starts from the same initial scene and selects additional views as needed.
Figure~\ref{fig:table_plot} compares system performance under visible and occluded conditions, while Tab.~\ref{tab:grasping_overall} reports grasp success pooled across all 150 trials.

With original instructions, the MLLM-based Point2Act and GraspMolmo achieve higher localization success on visible targets than LERF-TOGO and GraspSplats, although none of these baselines receives the event history.
Providing reasoned instructions improves localization and grasping success for every evaluated baseline.

As shown in Fig.~\ref{fig:table_plot}, our method localizes 43/50 visible and 88/100 occluded targets, exceeding the strongest baselines by 32 and 20 percentage points, respectively.
\textsc{Locate} restricts the pointing region to reduce distractor ambiguity, with Fig.~\ref{fig:reasoning}(B) illustrating its use for a small target part.
Grasping succeeds in 38/50 visible and 77/100 occluded trials, compared with 20/50 and 55/100 for the strongest baselines in each condition.
Across all 150 trials, our method achieves 76.7\% grasp success, compared with 50.0\% for the strongest baseline using reasoned instructions (Tab.~\ref{tab:grasping_overall}).
For occluded targets, our pipeline combines active view search with additional observation of local target geometry when needed for grasp validation.
\input{figures/active_perception_qualitative}

\subsection{Active Perception and Belief Ablations}
\label{sec:ap_belief_ablation}

We compare our method with an adapted implementation of the target-driven active-view method of Breyer~\etal~\cite{breyer_2022} on these four scenes (S1--S4 in Tab.~\ref{tab:active_perception_ablation}).

\subsubsection{Breyer~\etal}
Following the original method, we provide Breyer~\etal{} with the required ground-truth 3D bounding box of the target and retain the original TSDF-based ray-casting objective.
We replace VGN~\cite{breyer2020volumetric} with the AnyGrasp~\cite{anygrasp_2023} generator and planning stack shared with our method.
All methods stop at the first feasible target grasp, with target association based on the provided box for the baseline and the confirmed target region for ours and its ablations.
\input{figures/event_example}

\subsubsection{Belief Ablations}

We compare three variants in Tab.~\ref{tab:active_perception_ablation} while keeping candidate views, feasibility checks, target confirmation, and the grasp pipeline fixed.
\emph{Without belief weights} replaces the posterior-weighted objective in Eq.~\eqref{eq:view-score} with uniform visibility over target hypotheses not yet cleared by depth.
The posterior is still updated, but its probability mass no longer affects view ranking.
\emph{Without transmittance} sets $\sigma_t=0$ in Eq.~\eqref{eq:transmittance}, making $T_t=1$ and removing attenuation along sight lines through unobserved space.
\emph{Without event prior} replaces the history-derived initial belief with a uniform density while retaining subsequent belief updates and the view objective.

Our full method achieves 95\% grasp success with 2.20 views on average, versus 75\% success and 3.35 views for Breyer~\etal{} despite its privileged target box.
Although the baseline's box removes localization ambiguity, it neither reveals occluded target geometry nor guarantees grasp success.

Without belief weights, the mean view count is highest at 4.65 and grasp success rate falls to 80\%, suggesting that treating all remaining hypotheses equally wastes views on unlikely regions and reduces robustness.
%
% Without transmittance, the mean view count rises to 3.10, with S3 requiring 5.2 views versus 2.4 for ours, consistent with optimistic scoring of unknown-space sight lines.
Without transmittance, the mean view count rises to 3.10, with S3 requiring 5.2 views, consistent with optimistic scoring of unknown-space sight lines.
Without the event prior, grasp success matches that of the full method, but the mean view count rises from 2.20 to 4.00, showing the cost of searching without history-derived spatial guidance.
Fig.~\ref{fig:active_perception_qualitative} illustrates these differences in search behavior: in this example, our method reaches a target-revealing viewpoint directly, while the baseline explores additional viewpoints.

\subsection{Qualitative Video Reasoning on Egocentric Videos}
\label{sec:egocentric_reasoning}

To examine video reasoning beyond our recorded robot scenes, we apply the same pipeline to selected egocentric clips from EgoDex~\cite{egodex_2026} and EPIC-KITCHENS~\cite{epic_2020}.
Without dataset-specific prompt changes, the pipeline selects targets referred to by the order of plate placement or washing and produces 2D points on the requested objects in the final frames, as shown in Fig.~\ref{fig:reasoner}.
In the EPIC-KITCHENS example, which includes large camera viewpoint changes, \textsc{Locate} proposes an inaccurate region, but \textsc{Point} identifies the carrot within the expanded crop.
These examples illustrate use beyond our capture setup while showing that region estimation in the final frame can remain imprecise.

%% file: figures/table_plot.tex
\begin{figure*}[t]
    \centering
    \includegraphics[width=1.0\textwidth]{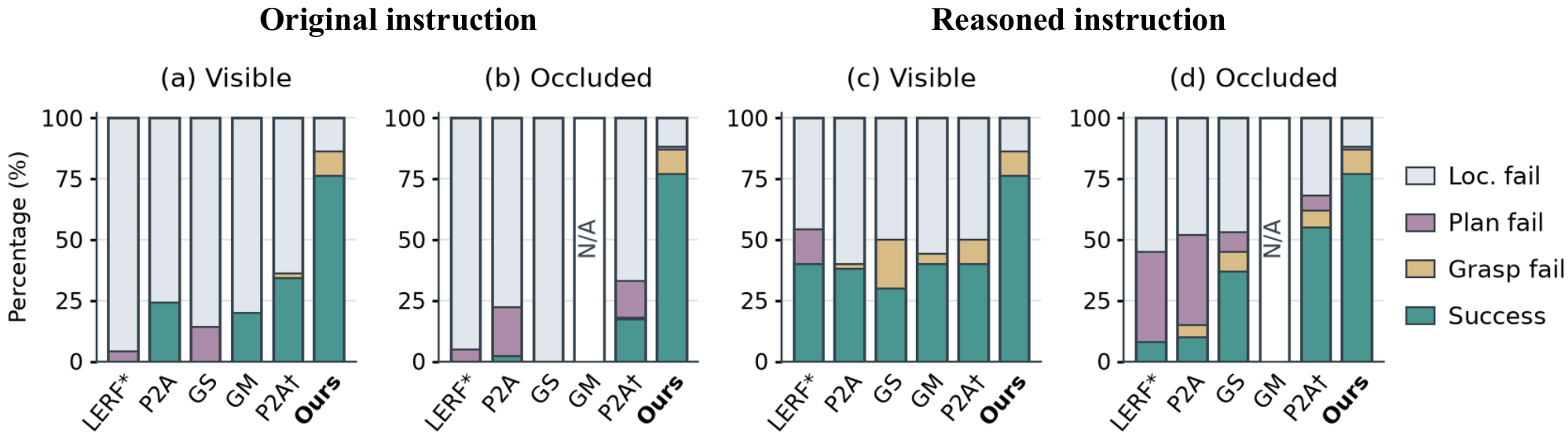}
    \caption{\textbf{Real-world grasping performance.}
    Targets are visible or occluded in the initial wrist-camera view, with 50 and 100 trials per method, respectively.
    Reasoned instructions are grounding queries generated by Qwen3-VL-8B~\cite{qwen3_2025} from the history video and original instruction, shared across baselines while retaining their native grounding backbones.
    LERF*, P2A, GS, and GM denote LERF-TOGO~\cite{lerf_togo_2023}, Point2Act~\cite{point2act_2025}, GraspSplats~\cite{graspsplat_2025}, and GraspMolmo~\cite{graspmolmo_2025}, respectively.
    P2A$^\dagger$ uses RGB-D reconstruction.
    Colored segments indicate successful grasps or the first unsuccessful stage.
    Ours is repeated as a common full-system reference across instruction conditions.
    N/A denotes an unevaluated condition.}
    \label{fig:table_plot}
\end{figure*}

%% file: tables/grasping_overall.tex
\begin{table}[t]
\centering
\caption{Grasp success (\%) over 150 trials.}
\label{tab:grasping_overall}
\begingroup
\footnotesize
\setlength{\tabcolsep}{2pt}
\renewcommand{\arraystretch}{1}
\begin{tabular*}{\columnwidth}{@{\extracolsep{\fill}}lccccc@{}}
\toprule
Instruction & LERF* & P2A & GS & P2A$^\dagger$ & Ours \\
\midrule
Original & 0.0 & 9.3 & 0.0 & 22.7 & \multirow{2}{*}{\textbf{76.7}} \\
Reasoned & 18.7 & 19.3 & 34.7 & 50.0 & \\
\bottomrule
\end{tabular*}
\endgroup
\end{table}

%% file: tables/active_perception_ablation.tex
\begin{table}[t]
\centering
\caption{Real-world active perception results and belief ablations on four heavily occluded scenes (S1--S4), five trials each.
View counts are averaged over trials, including failures. Grasp success aggregates all 20 trials.}
\label{tab:active_perception_ablation}
\begingroup
\setlength{\tabcolsep}{1.5pt}
\renewcommand{\arraystretch}{1}
% Reported Scenes 1--4 correspond to internal scene IDs 12--15.
\begin{tabular*}{\columnwidth}{@{\extracolsep{\fill}}lcccccc@{}}
\toprule
\multirow{2}{*}{Method} & \multicolumn{4}{c}{Scene} & \multicolumn{2}{c}{Overall} \\
\cmidrule(lr){2-5} \cmidrule(l){6-7}
& S1 & S2 & S3 & S4 & Views $\downarrow$ & Grasp Success $\uparrow$ \\
\midrule
Breyer~\etal~\cite{breyer_2022} & 4.2 & 4.2 & 2.2 & 2.8 & 3.35 & 15/20 (75\%) \\
\midrule
Ours w/o belief weights & 4.0 & 4.0 & 5.6 & 5.0 & 4.65 & 16/20 (80\%) \\
Ours w/o transmittance & 2.8 & 2.0 & 5.2 & 2.4 & 3.10 & 18/20 (90\%) \\
Ours w/o event prior & 3.8 & 4.0 & 4.2 & 4.0 & 4.00 & 19/20 (95\%) \\
\textbf{Ours} & 2.4 & 2.0 & 2.4 & 2.0 & \textbf{2.20} & \textbf{19/20 (95\%)} \\
\bottomrule
\end{tabular*}
\endgroup
\end{table}

%% file: figures/active_perception_qualitative.tex
\begin{figure}[t!]
    \centering
    \includegraphics[width=0.95\columnwidth]{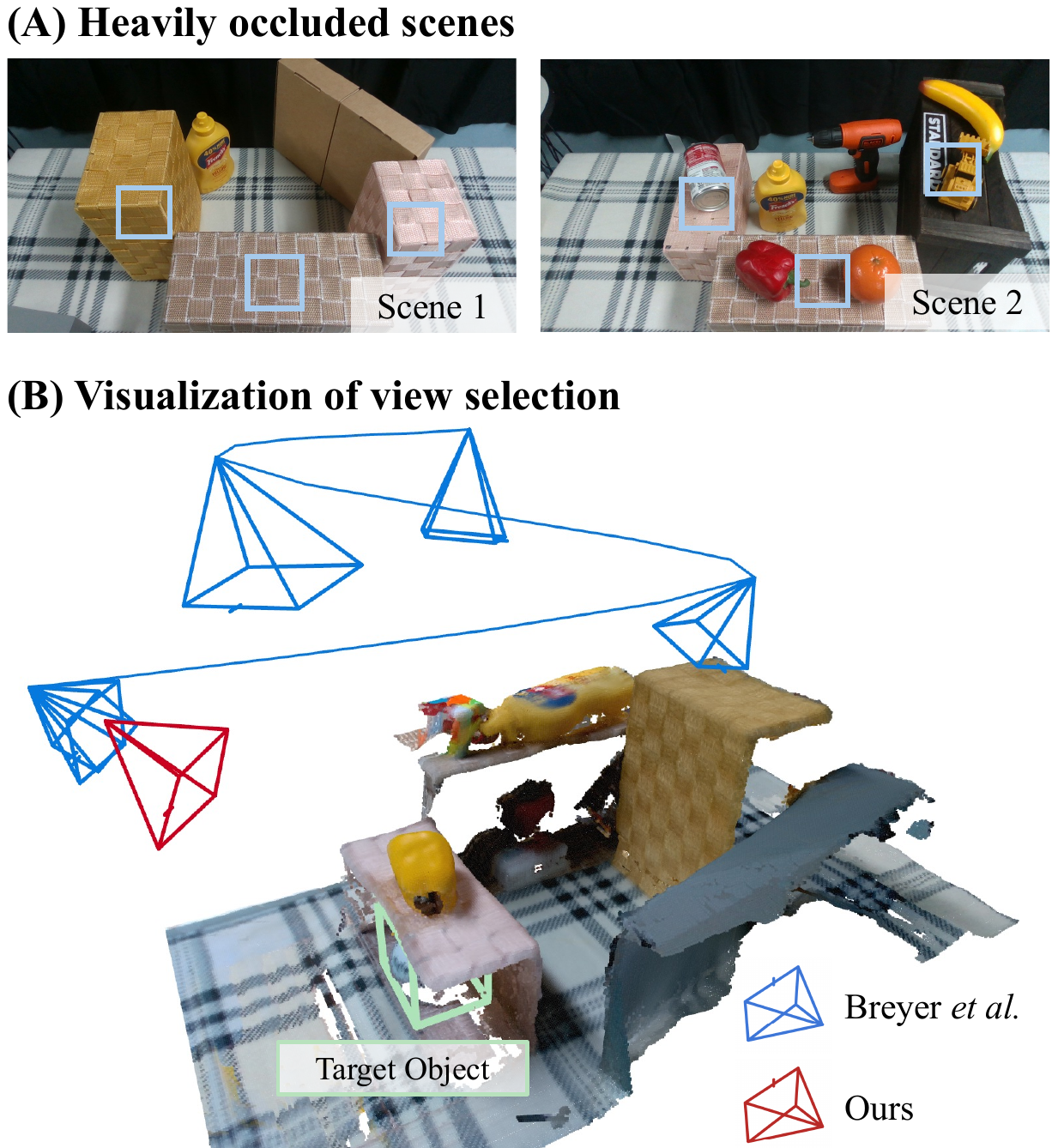}
    \caption{\textbf{Real-world active perception.}
    (A) Examples of heavily occluded scenes in which the queried targets are not visible from a fixed bird's-eye view. Boxed regions indicate locations where the object can be hidden.
    (B) Camera viewpoints acquired during target search.
    %Frustum colors denote \textcolor[RGB]{0,112,240}{\rule{1.2ex}{1.2ex}}~Breyer~\etal, \textcolor[RGB]{188,18,40}{\rule{1.2ex}{1.2ex}}~Ours, \textcolor[RGB]{215,108,13}{\rule{1.2ex}{1.2ex}}~w/o event prior, \textcolor[RGB]{220,179,25}{\rule{1.2ex}{1.2ex}}~w/o belief weights, and \textcolor[RGB]{96, 140, 16}{\rule{1.2ex}{1.2ex}}~w/o transmittance.
    In this example, our method reveals the target with just one additional view, whereas Breyer~\etal{} requires exploring multiple additional viewpoints.
    }
    \label{fig:active_perception_qualitative}
\end{figure}

%% file: figures/event_example.tex
\begin{figure}[t]
    \centering
    \includegraphics[width=0.9\columnwidth]{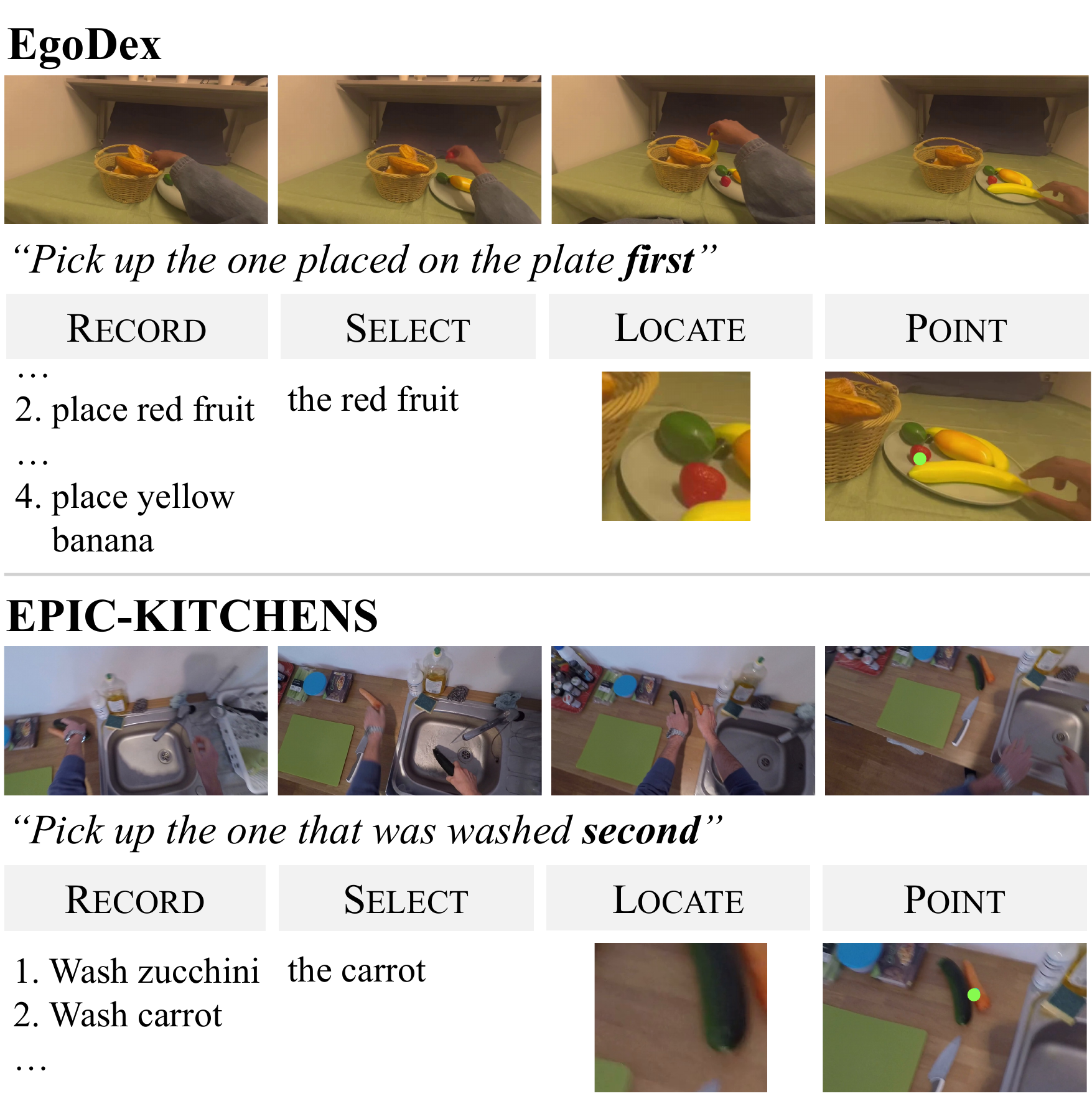}
    \caption{\textbf{Video reasoning examples on EgoDex~\cite{egodex_2026} and EPIC-KITCHENS~\cite{epic_2020}.}
    Each example pairs sampled event history and an instruction with target-related excerpts from the pipeline stages.
    Image panels show expanded proposal crops and enlarged final-frame regions, with predicted points.}
    \label{fig:reasoner}
    % \vspace{-2mm}
\end{figure}

%% file: sections/6_conclusion.tex
\section{Conclusion}
\label{sec:conclusion}
%After identifying a target from past events, a robot must still obtain the observations needed to grasp it under occlusion.
%We presented a zero-shot grasping system that connects the target's 3D observation history recovered from video to search in the current scene.
%The system combines a belief built from past target locations and current scene geometry with visibility estimates to select viewpoints for target localization and grasping.
%Our real-robot evaluation, with targets stationary during search, showed higher localization and grasping success than the evaluated baselines under both initially visible and occluded conditions.
%These results support using interaction history as spatial evidence that connects target identification to active observation.
To fulfill requests referring to a person's prior interactions, a robot must infer the intended target from the event history, then ground and grasp it in the current scene.
We present \Ours{}, a zero-shot system that connects video reasoning with active view selection for event-referential grasping.
The system identifies the requested object or part from the event history and, when the target is occluded, combines the recovered event prior with current scene geometry to obtain the observations needed for grasping.
Real-robot experiments demonstrate higher grasp success than the baselines for both visible and occluded targets.
Additional experiments under heavy occlusion show that the event prior reduce the number of observations while achieving a higher grasp success rate than the active-perception baseline.
Together, these results demonstrate the value of event history for both target identification and spatial search. 
Our formulation assumes that the target remains stationary during search. Extending the system to interactions in which objects continue to move remains future work.